\documentclass{article} 
\usepackage{iclr2024_conference,times}

\usepackage{amsmath,amsfonts,bm}

\def\eqref#1{equation~\ref{#1}}

\def\1{\bm{1}}

\DeclareMathAlphabet{\mathsfit}{\encodingdefault}{\sfdefault}{m}{sl}
\SetMathAlphabet{\mathsfit}{bold}{\encodingdefault}{\sfdefault}{bx}{n}

\usepackage{hyperref}
\usepackage{url}
\usepackage{microtype}
\usepackage{graphicx}
\usepackage{array}
\newcolumntype{?}{!{\vrule width 2pt}}
\usepackage{subfigure}
\usepackage{booktabs} 
\usepackage{algorithm,amsmath}
\usepackage{algpseudocode}
\usepackage{float}
\usepackage{caption}
\usepackage{bbm}
\usepackage{tikz}
\usepackage{multirow}
\usepackage{ctable}
\usepackage{courier}
\usepackage{dblfnote}
\usepackage{graphicx}

\usepackage{adjustbox}
\usepackage{graphics}

\usepackage{amsmath}
\usepackage{amssymb}
\usepackage{mathtools}
\usepackage{amsthm}

\usepackage[capitalize,noabbrev]{cleveref}

\title{Efficient Causal Graph Discovery Using Large Language Models}

\author{Thomas Jiralerspong * \\
Mila, Universit\'{e} de Montr\'{e}al\\
\texttt{thomas.jiralerspong@mila.quebec} \\
\And
Xiaoyin Chen * \\
Mila, Universit\'{e} de Montr\'{e}al\\
\texttt{xiaoyin.chen@mila.quebec} \\
\And
Yash More \\
Mila, McGill University\\
\And
Vedant Shah \\
Mila, Universit\'{e} de Montr\'{e}al\\
\And
Yoshua Bengio \\
Mila, Universit\'{e} de Montr\'{e}al\\
}

\def\thickhline{\noalign{\hrule height1.5pt}}
\iclrfinalcopy

\newcommand\blfootnote[1]{%
  \begingroup
  \renewcommand\thefootnote{}\footnote{#1}%
  \addtocounter{footnote}{-1}%
  \endgroup
}

\begin{document}
\blfootnote{$^*$Equal Contribution}
\maketitle

\begin{abstract}

We propose a novel framework that leverages LLMs for full causal graph discovery. While previous LLM-based methods have used a pairwise query approach, this requires a quadratic number of queries which quickly becomes impractical for larger causal graphs. In contrast, the proposed framework uses a breadth-first search (BFS) approach which allows it to use only a linear number of queries. We also show that the proposed method can easily incorporate observational data when available, to improve performance. In addition to being more time and data-efficient, the proposed framework achieves state-of-the-art results on real-world causal graphs of varying sizes. The results demonstrate the effectiveness and efficiency of the proposed method in discovering causal relationships, showcasing its potential for broad applicability in causal graph discovery tasks across different domains. 
\end{abstract}

\section{Introduction}
Recent advances in large language models (LLMs) have significantly improved AI capabilities \cite{openai2023gpt4}. LLMs have access to a vast amount of training data, allowing them to acquire a substantial amount of knowledge, ranging from common-sense to specific domains, such as math and science. Moreover, recent papers have argued that complex behaviors, such as writing code, generating long stories, and even some reasoning capabilities, can emerge from such large-scale training \cite{wei2022chainofthought, rozière2023code, zhao2023large, yao2023tree}. Impressively, LLMs can use their knowledge to generate plausible explanations given observations and even reason on counterfactuals \citep{wang2023pinto, li2023large,kıcıman2023causal}. Given these emergent abilities to understand and learn relationships from textual corpora, a natural question arises: \textit{Can LLMs reason about causal relationships?}

Causal discovery involves identifying the causal relationships between different variables \cite{Pearl09,causal}. This process usually results in a Directed Acyclic Graph (DAG), with edges representing the existence and direction of causal influences. This causal graph represents the relationships between variables which govern the data generation process. The causal graph also allows us to determine how changes in one variable might influence others, and serves as the foundation for further analysis related to specific tasks, including effect inference, prediction, or attribution. Ensuring the correctness of this causal graph is essential for the reliability of any subsequent analysis.

Traditional causal discovery frameworks \cite{bello2023dagma,zheng2018dags,Meek2023} rely on numerical observations and use statistical or continuous optimization based methods to unravel confounding and causal factors. While such rigorous numerical and abstract reasoning abilities remain challenges for LLMs \citep{gendron2023large,xu2023llms,imani2023mathprompter}, recent works have shown that LLMs are capable of leveraging their internal knowledge to perform commonsense and simple logical as well as causal reasoning \cite{wei2022chainofthought, zhao2023large, yao2023tree, kıcıman2023causal}. Inspired by these results, the proposed framework utilizes LLMs to respond to causal queries solely based on metadata, such as variable names and descriptions, without accessing the numerical observations. This is similar to the metadata and knowledge-based reasoning that is done by human domain experts when they construct causal graphs.

Previous works \cite{kıcıman2023causal,choi2022lmpriors,long2023large} have shown that LLMs are capable of inferring pairwise causal relationships given only metadata. However, extending this pairwise framework to full graph discovery presents additional challenges. One straightforward approach which previous works use is to perform pairwise queries over all possible pairs of variables. However, this method is not scalable since it requires $n \choose 2$ total queries where $n$ is the number of variables. Thus, the number of queries required grows quadratically with the number of causal variables. As a result, the experiments in these works are limited to very small causal graphs.



\begin{figure*}[t]
\begin{center}
\includegraphics[width=1.0\textwidth,trim={0 0 0 0cm},clip]{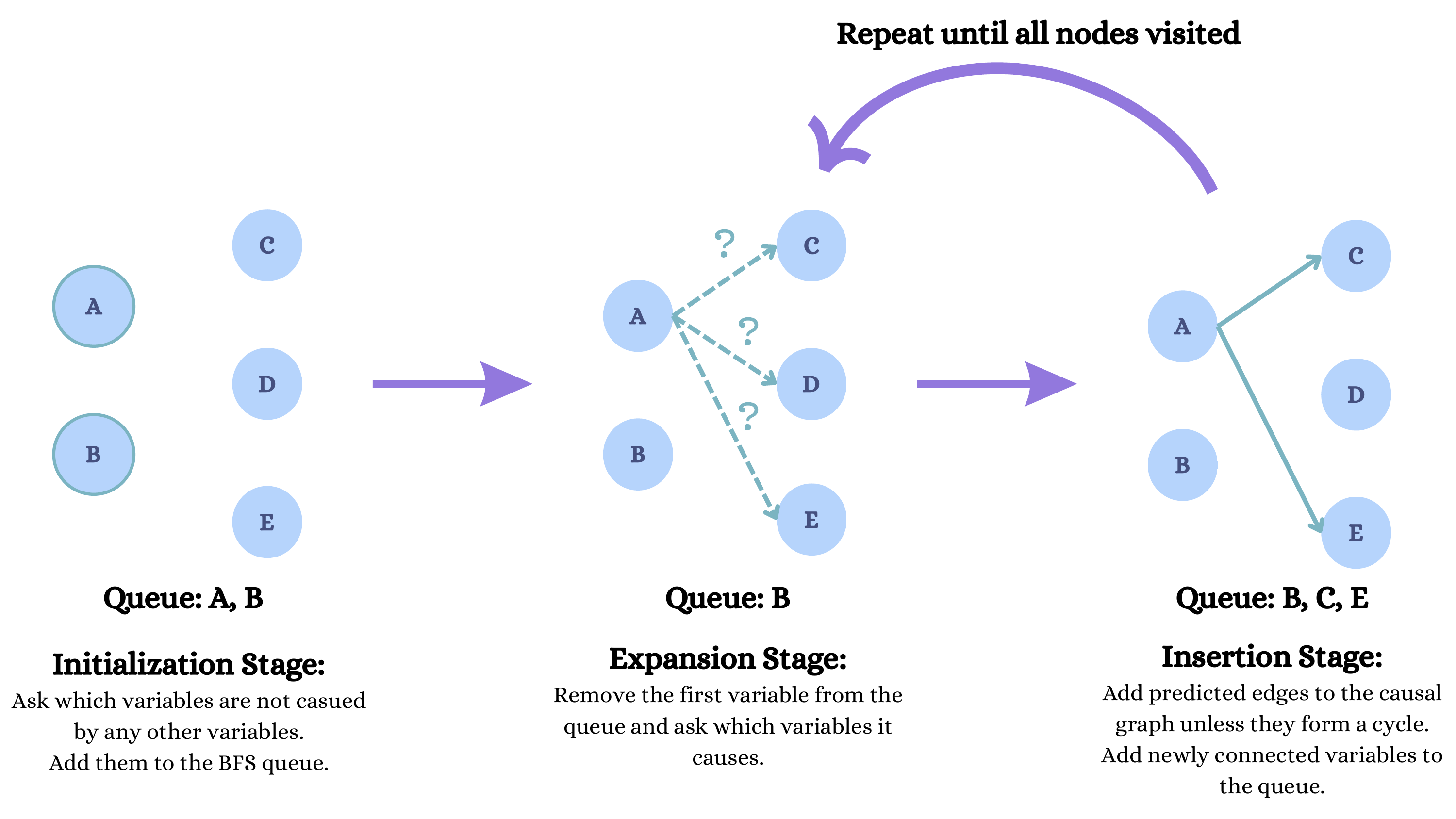}
\end{center}
\caption{Proposed framework for full graph discovery with LLMs}
\label{fig:framework}
\end{figure*}

To address the above problems, we propose a novel framework to discover full causal graphs with the internal knowledge of LLMs that guarantees the DAG constraint and avoids performing all pairwise queries (Figure \ref{fig:framework}). The key insight is that we leverage the DAG property of the resulting graph and construct the graph through a \textit{breadth-first search (BFS)}. Intuitively, node expansions in BFS correspond to prompting a LLM to select the variables caused by the current node. To efficiently construct causal graphs, we iteratively expand nodes with a LLM in BFS order and add the discovered edges. The proposed method achieves $O(n)$ query complexity compared to the $O(n^2)$ query complexity of the pairwise method. 

To the best of our knowledge, we present the first LLM-based causal graph discovery method applicable to larger graphs. The proposed method is much more efficient than previous LLM-based methods, and does not require observational data (unlike numerical methods), while still achieving state-of-the-art or competitive performance on three graphs of varying sizes. In addition, we show how observational data can easily be incorporated when available to improve the proposed method's performance. 

\section{Related Work}
\label{sec:related_works}

Causal graph discovery has been extensively studied in many settings \cite{zanga2023survey}. Until recently, most proposed methods were statistical methods that used observational data to infer causal relationships between variables. Some examples of these are Greedy Equivalence Search (GES) \cite{Meek2023}, the Peter-Clark Algorithm (PC) \cite{PC}, Non-combinatorial Optimization via Trace Exponential and Augmented lagrangian for Structure learning (NOTEARS) \cite{zheng2018dags}, and Directed Acyclic Graphs via M-matrices for Acyclicity (DAGMA) \cite{bello2023dagma}. Recently, many works have proposed to use LLMs for causal graph discovery \cite{kıcıman2023causal,choi2022lmpriors,long2023large}. These LLM-based methods do not rely on observational data. Instead, they utilize metadata associated with the input variables, specifically variable descriptions in natural language. This approach closely resembles how human experts construct causal graphs using their domain knowledge. 

\subsection{Pairwise Method}
\label{pairwise}
All of these works use pairwise queries to infer the causal relationship between 2 variables at a time. Given variables $A$ and $B$, there are three possible causal relationships, $A \rightarrow B$, $A \leftarrow B$, and no relationship. This query can be formulated as a question in natural language with the following prompt: 
\begin{verbatim}
<A>: <Description A>
<B>: <Description B>

Given the above information, which of the following is the most 
likely:
A. Changing <A> causes a change in <B>
B. Changing <B> causes a change in <A>
C. There is no causal relationship between <A> and <B>
\end{verbatim}

To extend to full graph discovery, \citet{kıcıman2023causal} propose to perform pairwise queries over all possible pairs. However, this method is not scalable since it requires $n \choose 2$ $ \in O(n^2)$ total queries, where $n$ is the number of variables, so they are only able to perform experiments on a small causal graph with 12 variables.

\section{Methods}
\subsection{Main Method}
\label{bfs}
We now propose a framework for efficient full causal graph discovery using a LLM. Instead of querying the LLM for each pair of nodes, we instruct the LLM to identify all potential effects of a single variable in one completion. This approach reduces the amount of queries required from $O(n^2)$ to $O(n)$ since visiting each node once is sufficient to determine all edges originating from that node.\\
With this new prompting method, we can treat each query as a node expansion in a BFS algorithm and gradually construct the causal graph by traversing it using BFS. To perform the BFS, we also need to establish the order in which nodes are explored. In this regard, we draw inspiration from the DAG property of causal graphs and utilize their topological orders for BFS. Specifically, the proposed framework consists of three stages: 1) an \textbf{Initialization stage} wherein the LLM is prompted to identify variables that are not causally affected by any other variables in the graph, 2) an \textbf{Expansion stage} wherein the LLM is instructed to find variables caused by the current node, and 3) an \textbf{Insertion stage} where the variables proposed by the LLM are added to the BFS queue and the edges proposed by the LLM are checked to make sure they do not cause cycles before being inserted. We start with the initialization stage, then alternate between the expansion and insertion stages until all nodes have been visited. The pseudocode for the proposed algorithm is shown in Algorithm~\ref{llm_bfs}.\\
\textbf{The initialization stage} determines variables that are not caused by any other variables. These selected variables are added to the the BFS queue to be used later as starting points for the BFS. 
We use the following \textit{initialization prompt} to provide the LLM with descriptions of all variables and ask the model to select independent variables. Note that the first paragraph of this prompt is task-specific. Here we use the prompt for experiments on the Neuropathic Pain Dataset (see section \ref{experimental-setup}) as an example:
\begin{verbatim}
You are a helpful assistant to a neuropathic pain diagnosis expert. 
The following factors are key variables related to neuropathic pain
diagnosis which have various causal effects on each other. Our goal 
is to construct a causal graph between these variables.

<A>: Description of variable A
<B>: Description of variable B
...

Now you are going to use the data to construct a causal graph. You 
will start with identifying the variable(s) that are unaffected by 
any other variables. Think step by step. Then, provide your final 
answer (variable names only) within the tags <Answer>...</Answer>. 
\end{verbatim}

\textbf{The expansion stage} expands each node in the search tree by prompting the LLM to select all variables that are caused by the current node. Given a queue of nodes that have not been visited yet (initialized by the initialization stage), we visit the nodes in a first-in-first-out order, i.e., the BFS order. We use the following \textit{expansion prompt} to provide the LLM with the current graph structure (i.e. the causal relationships already predicted) and let the model predict a list of affected variables:
\begin{verbatim}

Given <Independent Variables> is(are) not affected by any 
other variable and the following causal relationships:

A causes B, C, D
C causes D, E
…

Select the variables that are caused by <Currently Visited Node>.
Think step by step. Then, provide your final answer (variable names 
only) within the tags <Answer>...</Answer>.
\end{verbatim}
In the \textbf{insertion stage}, for each variable predicted by the model, we first add it to the queue of nodes to be visited. Next, for each edge from the current node to its expanded children, we perform a cycle check before adding this edge to the predicted causal graph. This check is necessary because, although the LLM provides the current graph structure, we observe that it often generates edges that violate the DAG constraint. To check for cycles, we perform a depth-first search in the predicted graph with the edge to be added. If adding the edge creates a cycle, we do not add it, thus guaranteeing that the resulting graph is a DAG. \\
We alternate between the expansion and insertion stages until all nodes have been visited. The number of queries required for the proposed  algorithm is $O(n)$, as BFS only visits each node once. This is in contrast to the pairwise method (Sec.~\ref{pairwise}), which requires $O(n^2)$ queries. \\

\subsection{Adding Statistics to the Prompt}
We also experiment with adding relevant statistics to the prompt in the form of the Pearson correlation coefficients between the variables involved in the prompt. Even though correlation does not imply causation, we hypothesize that adding such statistics to the prompt helps the LLM by giving it statistical relationships which it can use when its internal knowledge is not enough to determine the causal relationship between 2 variables.

For the proposed method, this is done by adding the following text to the \textit{expansion stage} prompt after the list of causal relationships:
\begin{verbatim}
Additionally, the Pearson Correlation Coefficients between 
<Currently Visited Node> and the other variables are as follows:
<A>: <Pearson Correlation Coefficient between <Currently 
Visited Node> and A>
<B>: <Pearson Correlation Coefficient between <Currently 
Visited Node> and B>
...
\end{verbatim}

For the pairwise method (Sec. \ref{pairwise}), the following text is added to the prompt after the name and description of the variables:
\begin{verbatim}
Additionally, the Pearson Correlation Coefficient between <A> and 
<B> is <Correlation Coefficient>.
\end{verbatim}

Lastly, it is important to mention that the search process takes place within a single multi-turn chat environment. This enables the LLM to base each prompt on the preceding chat history.

\begin{algorithm}[htbp]
\caption{BFS with LLMs}
\label{llm_bfs}
\begin{algorithmic}
\Require LM $p_\theta$, descriptions of variables $X$, initial variable selector $I()$, expansion generator $E()$, cycle checker $CheckCycle()$
\State $G \gets \{\}$ \Comment{Create an empty graph to store the result.}
\State $frontier,visited \gets I(p_\theta,X)$ \Comment{With initialization prompt.}
\While{$frontier$ is not empty}
\State $toVisit \gets frontier\left[0\right]$ 
\State $frontier.remove(toVisit)$
\State $visited.add(toVisit)$
\For{$node$ in $E(p_\theta, G)$} \Comment{Expand with expansion prompt.}
\If{not $CheckCycle(G,toVisit,node)$} 
\State \Comment{Check if adding $toVisit \rightarrow node$ will create cycle.}
\State$G.add((toVisit,node))$
\EndIf

\If{$node$ not in $frontier \cup visited$} 
\State$frontier.add(node)$
\EndIf

\EndFor
\EndWhile
\State return $G$
\end{algorithmic}
\end{algorithm}

\section{Experiments and Results}

\subsection{Experimental Setup}
\label{experimental-setup}

We use the GPT-4 API \cite{openai2023gpt4} for all of our experiments. We evaluate the proposed approach on three causal graphs of varying sizes:

\begin{enumerate}
    \item \textbf{Asia} \cite{asia_dataset}. The Asia causal graph is a small causal graph that models lung cancer in patients having potentially recently visited Asia. The ground-truth graph has 8 nodes and 8 edges. Some examples of variables in the graph are \textit{Visit to Asia}, \textit{Smoking}, \textit{Bronchitis}, and \textit{Lung Cancer}.
    \item \textbf{Child} \cite{child_dataset}. The Child causal graph is a medium-sized causal graph which models congenital heart disease in newborn babies. The ground-truth graph has 20 nodes and 25 edges. Some examples of variables in the graph are \textit{Birth Asphyxia}, \textit{Lung Flow}, and \textit{Chest XRay}.
    \item \textbf{Neuropathic Pain} \cite{tu2019neuropathic} The Neuropathic Pain causal graph is a very large causal graph consisting of 221 nodes and 770 edges. It contains relationships between different nerves and the associated symptoms that patients express. It  mainly consists of \textit{symptom diagnoses}, \textit{pattern diagnoses}, and \textit{pathophysiological diagnoses}. The \textit{symptom diagnoses} describe the discomfort of patients. The \textit{pattern diagnoses} identify nerve roots that lead to symptom patterns. The \textit{pathophysiological diagnoses }identify discoligamentous injuries. 
\end{enumerate}

For the Neuropathic Pain causal graph, we use the same prompts as described in Section \ref{bfs}. We use similar prompts with appropriate task specific modifications for the other two datasets.

We get the Asia and Child causal graphs and data from the \texttt{bnlearn} package \cite{bnlearn}, and their descriptions from \citet{long2023causal}. We get the Neuropathic Pain causal graph data and descriptions from \citet{tu2019neuropathic}.

We compare the proposed method, as well as the proposed method plus observational statistics from 1000 and 10000  samples, against the numerical methods of Greedy Equivalence Search (GES) \cite{Meek2023}, the Peter-Clark Algorithm (PC) \cite{PC}, Non-combinatorial Optimization via Trace Exponential and Augmented lagrangian for Structure learning (NOTEARS) \cite{zheng2018dags}, and Directed Acyclic Graphs via M-matrices for Acyclicity (DAGMA) \cite{bello2023dagma}, as well as against pairwise queries, and pairwise queries plus observational statistics from 1000 and 10000 samples. We use the implementations of GES and PC from the \texttt{causal-learn} package \cite{causallearn}, the implementation of NOTEARS from \citet{zheng2018dags}, and the implementation of DAGMA from \citet{bello2023dagma}. For the numerical methods, we conduct experiments in the low (100 samples), medium (1000 samples), and high (10000 samples) data regimes. 

For NOTEARS and DAGMA, we try $\lambda$ values in $\{0, 0.005, 0.1, 0.2\}$ and report the results for the best performing $\lambda$ value. 

For the LLM methods, we use the newest \textit{gpt-4-0125-preview} checkpoint. We try sampling temperatures in $\{0, 0.5, 0.7, 1.0\}$ and report the results for the best temperature setting. 

\subsection{Metrics}
Causal graph discovery can be viewed as a classification task where each pair of variables (A,B) must be classified as either \textit{A causes B}, \textit{B causes A}, or \textit{Neither}. We therefore report standard classification metrics such as accuracy, precision, recall, and F score.

Following \citet{kıcıman2023causal}, we also report the Normalized Hamming Distance (NHD). In a graph with $m$ nodes, the NHD is given by $\sum_{i,j=1}^m \frac{1}{m^2} \mathbbm{1}_{G_{ij} \neq G'_{ij}}$, i.e. the number of edges that are present in one graph and not the other, divided by the total number of all possible edges. One problem with the NHD is that since most graphs are relatively sparse, trivially outputting 0 predicted edges achieves a relatively good NHD. For example, for the Child graph, which has 20 nodes and 25 edges, if we output 0 predicted edges then we achieve a NHD of $ \frac{1}{(20^2)} 25 = 0.0625$, which is better than the NHD achieved by all the methods. Intuitively, predicting an edge is much riskier when using this metric, so algorithms that predict more edges are punished. To remedy this and take into account the number of predicted edges, again following \citet{kıcıman2023causal}, we compute the ratio between the NHD and the baseline NHD of an algorithm that outputs the same number of edges but with all of them being incorrect. The lower the ratio, the better the method does compared to the worst baseline that outputs the same number of edges. We report the NHD, number of predicted edges (NPE), baseline NHD, and ratio between NHD and baseline NHD for all algorithms.

As our main metrics for comparison, we use the ratio between NHD and baseline NHD as well as the F score since these metrics are the least biased by the number of predicted edges.

\begin{table}
\vspace{-0.3cm}
\begin{center}
\resizebox{\textwidth}{!}{
\begin{tabular}{c|c|cccc|cccc|cccc}
\toprule
                                    & & \multicolumn{4}{c}{ASIA (n=8, m=8)}   & \multicolumn{4}{c}{CHILD (n=20, m=25)} & \multicolumn{4}{c}{NEUROPATHIC  (n=221, m=770)}  \\
\midrule
\# of Samples & Method &  Acc. ($\uparrow$) & F Score ($\uparrow$) & NPE & Ratio ($\downarrow$) &  Acc. ($\uparrow$) & F Score ($\uparrow$) & NPE & Ratio ($\downarrow$) &  Acc. ($\uparrow$) & F Score ($\uparrow$) & NPE & Ratio ($\downarrow$) \\
\midrule
\multirow{6}{*}{100} & GES & 0.23 & 0.38 & 8 & 0.63 & 0.24 & 0.38 & 22 & 0.62 & - & - & - & - \\
& PC & 0.33 & 0.5 & 4 & 0.5 & 0.19 & 0.32 & 19 & 0.68 & 0.02 & 0.04 & 311 & 0.96 \\
& NOTEARS & 0.5 & 0.67 & 4 & 0.33 & 0.14 & 0.25 & 23 & 0.75 & 0.029 & 0.06 & 76 & 0.94 \\
& DAGMA & 0.22 & 0.36 & 3 & 0.64 & 0.14 & 0.24 & 41 & 0.76 & 0.03 & 0.05 & 77 & 0.95 \\
& Pairwise & 0.33 & 0.5 & 20 & 0.5 & 0.3 & 0.46 & 27 & 0.54 & - & - & - & - \\
& Ours & \textbf{0.88} & \textbf{0.93} & 7 & \textbf{0.067} & 0.3 & 0.46 & 27 & 0.54 & \textbf{0.22} & \textbf{0.35} & 331 & \textbf{0.64} \\
\midrule
\multirow{6}{*}{1000} & GES & 0.67 & 0.8 & 7 & 0.2 & 0.31 & 0.44 & 34 & 0.53 & - & - & - & - \\
& PC & 0.5 & 0.67 & 7 & 0.33 & 0.29 & 0.45 & 37 & 0.55 & 0.04 & 0.08 & 559 & 0.92 \\
& NOTEARS & 0.44 & 0.62 & 5 & 0.38 & 0.2 & 0.33 & 17 & 0.67 & 0.02 & 0.04 & 36 & 0.96 \\
& DAGMA & 0.5 & 0.67 & 4 & 0.33 & 0.24 & 0.39 & 21 & 0.61 & 0.03 & 0.057 & 316 & 0.94 \\
& Pairwise & 0.38 & 0.54 & 14 & 0.45 & 0.4 & 0.57 & 24 & 0.43 & - & - & - & - \\
& Ours & 0.8& 0.89 & 10 & 0.11 & 0.4 & 0.57 & 24 & 0.43 & - & - & - & - \\
\midrule
\multirow{6}{*}{10000} & GES & 0.7 & 0.82 & 9 & 0.18 & 0.42 & 0.58 & 47 & 0.42 & - & - & - & - \\
& PC & 0.55 & 0.71 & 9 & 0.29 & 0.26 & 0.42 & 47 & 0.58 & - & - & - & - \\
& NOTEARS & 0.44 & 0.62 & 5 & 0.38 & 0.19 & 0.32 & 19 & 0.68 & 0.03  & 0.058 & 53 & 0.94 \\
& DAGMA & 0.5 & 0.67 & 4 & 0.33 & 0.22 & 0.36 & 20 & 0.64 & 0.033 & 0.063 & 214 & 0.94 \\
& Pairwise & 0.47 & 0.64 & 17 & 0.36 & 0.14 & 0.25 & 86 & 0.75 & - & - & - & - \\
& Ours & 0.8 & 0.89 & 10 & 0.11 & \textbf{0.46} & \textbf{0.63} & 25 & \textbf{0.37} & - & - & - & - \\
\bottomrule
\end{tabular}
}
\end{center}
\vspace{-0.3cm}
\caption{Results on the Asia (8 nodes, 8 edges), Child (20 nodes, 25 edges), and Neuropathic (221 nodes, 770 edges) causal graphs. Our method obtains state-of-the-art performance on all 3 causal graphs. More results can be found in the appendix.}
\label{all_result}

\end{table}

\subsection{Results}

Tables \ref{all_result} shows the results of our experiments on the Asia causal graph, the Child causal graph, and the Neuropathic Pain causal graph respectively. We bold the best result in the F Score column and NHD Ratio column since those are the main metrics we are looking at for comparison.

On the \textbf{Asia} causal graph, the proposed method without observational statistics performs the best of all the methods by far, achieving an F score of 0.93 and a NHD ratio of 0.067. Of the numerical methods, in the low data regime, NOTEARS performs the best, while GES performs the best in the medium and high data regimes. Interestingly, adding observational statistics to the prompt does not improve the performance of the proposed method for this graph, potentially because it is small enough for the LLM to be able to use its common sense knowledge to discover the causal relationships between all of its variables, so the addition of observational statistics is unhelpful and only adds unnecessary noise to the prompt.

On the \textbf{Child} causal graph, the proposed method with observational statistics from 10000 samples performs the best out of all the methods, with an F score of 0.63 and a NHD Ratio of 0.37. The proposed method with observational statistics from 1000 samples performs almost exactly as well as the best baseline (GES with 10000 samples) while using 10x less data. Finally, the proposed method without observational statistics is beaten only by GES with 10000 samples and performs as well as the next best baseline (GES with 1000 samples) while using no observational data. These results also showcase how incorporating observational statistics into the LLM's prompt can significantly improve performance.

On the \textbf{Neuropathic Pain} causal graph, running pairwise queries would require $221 \choose 2$ $= 24310$ queries, which is impractical. Similarly, GES requires considering all 24310 possible edges at each step of its search, which is intractable. We thus omit the results for both GES and pairwise queries on the Neuropathic Pain graph. In addition, PC runs out of memory when run on the Neuropathic Pain graph with 10000 samples, so we also omit those results from the table. 

On this much larger graph, we see that even the methods that are able to run fail catastrophically, barely outperforming the baseline which predicts the same number of edges but all wrong (as seen by the NHD Ratios which are all above 0.9). In contrast, the proposed method achieves reasonable performance with an F Score of 0.351 and a NHD Ratio of 0.643, thus emerging as the only method which is able to handle larger graphs such as this one.

Since the proposed method without observational data already outperforms the other methods significantly on the Neuropathic Pain causal graph and the number of queries required for this large graph is quite significant even considering the efficiency of The proposed method, we choose not to run the proposed method with observational statistics on this graph due to budget constraints.

\section{Limitations}
The proposed method has the same limitations as any method that relies on LLMs.

First, since LLMs rely on the knowledge from their training data to synthesize the causal relationships, the proposed approach only works on real-world data. Second, due to their black-box nature, the generations made by the LLM lack attribution or grounding, limiting transparency. Finally, the performance of the proposed method is dependent on the quality of the LLM used.

\section{Conclusion}
In this work, we introduce a novel method for causal graph discovery which leverages a breadth-first search approach in conjunction with Large Language Models. In contrast to statistical approaches, the proposed method eliminates the need for observational data and extra training time, ensuring quick and easy applicability across diverse problem domains. When observational data is available, we show that it can be incorporated into the proposed method to further improve performance. Compared to existing LLM-based methods, the proposed approach achieves superior efficiency by scaling the number of queries linearly with the number of variables, rather than quadratically. Furthermore, the proposed method achieves state-of-the-art performance on three causal graphs of varying sizes, and is the only method tested to achieve reasonable performance on the large Neuropathic Pain causal graph. The proposed method thus emerges as a more efficient, flexible, and effective solution to causal graph discovery.

\section{Future Work}
One avenue that future work should aim to explore is further combining the common sense knowledge of LLMs with information from observational data by employing hybrid statistical/LLM approaches. Potential strategies include using the LLM-generated causal graph as a prior for statistical methods (similar to \citet{choi2022lmpriors}), employing statistical methods to validate causal relationships suggested by the LLM against observational data, and incorporating different observational statistics into the LLM prompt.

Another avenue to explore is using advanced prompting strategies like Tree of Thoughts \citep{yao2023tree}, which are known to significantly improve performance in complex multi-step reasoning tasks like causal graph discovery.

A final avenue to explore is trying the proposed method with different LLMs, in order to provide insights into how the size and quality of the LLM influences the proposed method's performance. 

\section{Reproducibility Statement}
We provide code to reproduce all experiments in this \href{https://github.com/superkaiba/causal-llm-bfs.git}{repository}

\section{Acknowledgements}
We are grateful to  Dhanya Sridhar, Philippe Brouillard, and Marco Jiralerspong for their helpful ideas and suggestions.


\newpage
\appendix

\bibliography{iclr2024_conference}
\bibliographystyle{iclr2024_conference}

\appendix

\begin{table*}[h!]
\captionsetup{justification=centering}

\centering
\resizebox{\textwidth}{!}{
\begin{tabular}{lllllllllll}
\thickhline
Method &  Acc. $\uparrow$ & Prec. $\uparrow$ & Recall $\uparrow$ & F Score $\uparrow$ & NPE & NHD $\downarrow$& Ref. NHD & NHD Ratio $\downarrow$\\
\thickhline
\textbf{\textit{Numerical Methods}} & \multicolumn{1}{l}{}    & \\
\thickhline
\textit{100 Samples} & \multicolumn{1}{l}{}    & \\
\hline
GES & 0.23 & 0.38 & 0.38 & 0.38 & 8 & 0.28 & 0.44 & 0.63 \\
PC & 0.33 & 0.75 & 0.38 & 0.5 & 4  &  0.24 & 0.48 & 0.5 \\
NOTEARS ($\lambda=0.02$) & 0.5 & 1 & 0.5 & 0.67 & 4  & 0.11 & 0.33 & 0.33 \\
DAGMA ($\lambda=0.02$) & 0.22 & 0.67 & 0.25 & 0.36 & 3  &  0.28 & 0.44 & 0.64 \\

\hline
\textit{1000 Samples} & \multicolumn{1}{l}{}    & \\
\hline
GES & 0.67 & 0.86 & 0.75 & 0.8 & 7 & 0.06 & 0.31 & 0.2 \\
PC & 0.5 & 0.71 & 0.63 & 0.67 & 7 &   0.1 & 0.31 & 0.33 \\
NOTEARS ($\lambda=0.005$)& 0.44 & 0.8 & 0.5 & 0.62 & 5 &   0.10 & 0.27 & 0.38 \\
DAGMA ($\lambda=0.01$) & 0.5 & 1 & 0.5 & 0.67 & 4 &   0.11 & 0.33 & 0.33 \\

\hline
\textit{10000 Samples} & \multicolumn{1}{l}{}    & \\
\hline
GES & 0.7 & 0.78 & 0.88 & 0.82 & 9 & 0.06 & 0.35 & 0.18 \\
PC & 0.55 & 0.67 & 0.75 & 0.71 & 9  & 0.1 & 0.35 & 0.29 \\
NOTEARS ($\lambda=0.005$) & 0.44 & 0.8 & 0.5 & 0.62 & 5  & 0.14 & 0.36 & 0.38 \\
DAGMA ($\lambda=0.005$) & 0.5 & 1 & 0.5 & 0.67 & 4 & 0.11 & 0.33 & 0.33 \\

\thickhline

\textbf{\textit{LLM Methods}} & \multicolumn{1}{l}{}    & \\
\thickhline
Pairwise & 0.33 & 0.35 & 0.88 & 0.5 & 20   & 0.22 & 0.44 & 0.5 \\
Pairwise + 1000 samples & 0.38 & 0.43 & 0.75 & 0.54 & 14 & 0.20 & 0.45 & 0.45 \\
Pairwise + 10000 samples & 0.47 & 0.47 & 1 & 0.64 & 17 & 0.14 & 0.39 & 0.36 \\
Ours & 0.88 & 1 & 0.88 & \textbf{0.93} & 7 & 0.016 &  0.23 & \textbf{0.067} \\
Ours + 1000 samples & 0.8 & 0.8 & 1 & 0.89 & 10 & 0.031 &  0.28 & 0.11 \\
Ours + 10000 samples & 0.8 & 0.8 & 1 & 0.89 & 10 & 0.031 &  0.28 & 0.11 \\

\end{tabular}}
\caption{Results on the Asia causal graph (8 nodes, 8 edges). The proposed method with no observational statistics outperforms all other methods with a F score of 0.93 and a NHD ratio of 0.067.}
\label{table:asia}
\end{table*}

\begin{table*}[ht!]
\captionsetup{justification=centering}
\centering
\resizebox{\textwidth}{!}{
\begin{tabular}{lllllllllll}
\thickhline
Method &  Acc. $\uparrow$ & Prec. $\uparrow$ & Recall $\uparrow$ & F Score $\uparrow$ & NPE & NHD $\downarrow$& Ref. NHD & NHD Ratio $\downarrow$\\
\thickhline
\textbf{\textit{Numerical Methods}} & \multicolumn{1}{l}{}    & \\
\thickhline
\textit{100 Samples} & \multicolumn{1}{l}{}    & \\
\hline
GES & 0.24 & 0.41 & 0.36 & 0.38 & 22 & 0.11 & 0.18 & 0.62 \\
PC & 0.19 & 0.37 & 0.28 & 0.32 & 19& 0.12 & 0.17 & 0.68 \\
NOTEARS ($\lambda=0.01$)& 0.14 & 0.26 & 0.24 & 0.25 & 23 &  0.1 & 0.13 & 0.75 \\
DAGMA ($\lambda=0.005$) & 0.14 & 0.2 & 0.32 & 0.24 & 41 & 0.13 & 0.17 & 0.76 \\
\hline
\textit{1000 Samples} & \multicolumn{1}{l}{}    & \\
\hline
GES & 0.31 & 0.41 & 0.47 & 0.44 & 34 & 0.08 & 0.15 & 0.53 \\
PC & 0.29 & 0.38 & 0.56 & 0.45 & 37 &   0.085 & 0.155 & 0.55 \\
NOTEARS ($\lambda=0.005$)& 0.2 & 0.41 & 0.28 & 0.33 & 17 & 0.086 & 0.13 & 0.67 \\
DAGMA ($\lambda=0.005$) & 0.24 & 0.43 & 0.36 & 0.39 & 21 & 0.07 & 0.12 & 0.61 \\
\hline
\textit{10000 Samples} & \multicolumn{1}{l}{}    & \\
\hline
GES & 0.42 & 0.45 & 0.84 & 0.58 & 47 &  0.08 & 0.18 & 0.42 \\
PC & 0.26 & 0.32 & 0.6 & 0.42 & 47 &   0.101 & 0.18 & 0.58 \\
NOTEARS ($\lambda=0.005$)& 0.19 & 0.37 & 0.28 & 0.32 & 19 &  0.093 & 0.14 & 0.68 \\
DAGMA ($\lambda=0$) & 0.22 & 0.4 & 0.32 & 0.36 & 20 &   0.08 & 0.12 & 0.64 \\
\thickhline

\textbf{\textit{LLM Methods}} & \multicolumn{1}{l}{}    & \\
\thickhline
Pairwise & 0.13 & 0.16 & 0.4 & 0.23 & 61 & 0.165 & 0.215 & 0.76 \\
Pairwise + 1000 samples & 0.16 & 0.19 & 0.52 & 0.28 & 25 & 0.17 & 0.24 & 0.72 \\
Pairwise + 10000 samples & 0.14 & 0.16 & 0.56 & 0.25 & 86 & 0.21 & 0.28 & 0.75 \\
Ours & 0.3 & 0.44 & 0.48 & 0.46 & 27 & 0.07 &  0.13 & 0.54 \\
Ours + 1000 samples & 0.4 & 0.58 & 0.56 & 0.57 & 24  & 0.065 &  0.15 & 0.43 \\
Ours + 10000 samples & 0.46 & 0.59 & 0.68 & \textbf{0.63} & 25  & 0.055 &  0.15 & \textbf{0.37} \\

\end{tabular}}
\caption{Results on the Child causal graph (20 nodes, 25 edges). The proposed method with observational statistics from 10000 samples outperforms all other methods with a F score of 0.63 and a NHD ratio of 0.37.}
\label{table:child}
\end{table*}

\begin{table*}[ht!]
\captionsetup{justification=centering}
\centering
\resizebox{\textwidth}{!}{
\begin{tabular}{lllllllllll}
\thickhline
Method &  Acc. $\uparrow$ & Prec. $\uparrow$ & Recall $\uparrow$ & F Score $\uparrow$ & NPE & NHD $\downarrow$& Ref. NHD & NHD Ratio $\downarrow$\\
\thickhline
\textbf{\textit{Numerical Methods}} & \multicolumn{1}{l}{}    & \\
\thickhline
\textit{100 Samples} & \multicolumn{1}{l}{}    & \\
\hline
GES & N/A & N/A & N/A & N/A & N/A &  N/A & N/A & N/A \\
PC & 0.02 & 0.08 & 0.03 & 0.04 & 311 &  0.022 & 0.023 & 0.96 \\
NOTEARS ($\lambda=0.01$)& 0.029 & 0.32 & 0.03 & 0.06 & 76  & 0.1 & 0.11 & 0.94 \\
DAGMA ($\lambda=0.01$) & 0.03 & 0.3 & 0.03 & 0.05 & 77  & 0.10 & 0.11 & 0.95 \\
\hline
\textit{1000 Samples} & \multicolumn{1}{l}{}    & \\
\hline
GES & N/A & N/A & N/A & N/A & N/A & N/A  & N/A & N/A \\
PC & 0.04 & 0.09 & 0.07 & 0.08 & 559 &  0.025 & 0.027 & 0.92 \\
NOTEARS ($\lambda=0.01$)& 0.02 & 0.47 & 0.02 & 0.04 & 36 &   0.34 & 0.35 & 0.96 \\
DAGMA ($\lambda=0$) & 0.03 & 0.098 & 0.040 & 0.057 & 316 &  0.040 & 0.042 & 0.94 \\

\hline
\textit{10000 Samples} & \multicolumn{1}{l}{}    & \\
\hline
GES & N/A & N/A & N/A  & N/A & N/A  & N/A & N/A & N/A \\
PC & N/A & N/A & N/A  & N/A & N/A & N/A & N/A & N/A \\
NOTEARS ($\lambda=0.005$)& 0.03 & 0.45 & 0.031 & 0.058 & 53  & 0.19 & 0.20 & 0.94 \\
DAGMA ($\lambda=0$) & 0.033 & 0.14 & 0.040 & 0.063 & 214 & 0.059 & 0.063 & 0.94 \\

\thickhline

\textbf{\textit{LLM Methods}} & \multicolumn{1}{l}{}    & \\
\thickhline
Pairwise & N/A & N/A & N/A &  N/A & N/A & N/A & N/A & N/A \\
Ours & 0.217 & 0.583 & 0.251 & \textbf{0.351} & 331  & 0.014 & 0.022 & \textbf{0.643} \\

\end{tabular}}
\caption{Results on the Neuropathic Pain causal graph (221 nodes, 770 edges). We omit the results for GES and pairwise queries because they are intractable to use on a graph of this size. All methods except the proposed method fail catastrophically. The proposed method performs reasonably well with a F score of 0.351 and a NHD ratio of 0.643.}
\label{table:neuropathic}
\end{table*}

\end{document}